# Mechanistic Interpretability of Fine-Tuned Vision Transformers on Distorted Images: Decoding Attention Head Behavior for Transparent and Trustworthy AI

Nooshin Bahador

Krembil Research Institute, University Health Network, Toronto, Canada.

- [Access to Python package for Attention Maps Extraction](#)
- [Access to Python Package for Attention Head Analysis](#)
- [Access to Trained Vision Transformer on Synthetic Spectrograms](#)
- [Download Link for 100,000 Synthetic Spectrogram Dataset for Chirp Localization (Images & Corresponding Labels)](#)
- [Repository with PyTorch Implementation for Fine-Tuning the Vision Transformer](#)
- [Python package for generating synthetic chirp spectrogram (Images & Corresponding labels](#))

**Abstract**

Mechanistic interpretability improves the safety, reliability, and robustness of large AI models. This study examined individual attention heads in vision transformers (ViTs) fine-tuned on distorted 2D spectrogram images containing non-relevant content (axis labels, titles, color bars). By introducing extraneous features, the study analyzed how transformer components processed unrelated information, using mechanistic interpretability to debug issues and reveal insights into transformer architectures. Attention maps assessed head contributions across layers. Heads in early layers (1-3) showed minimal task impact with ablation increased MSE loss slightly ($\mu=0.11\%$, $\sigma=0.09\%$), indicating focus on less critical low level features. In contrast, deeper heads (e.g., layer 6) caused a threefold higher loss increase ($\mu=0.34\%$, $\sigma=0.02\%$), demonstrating greater task importance. Intermediate layers (6-11) exhibited monosemantic behavior, attending exclusively to chirp regions. Some early heads (1-4) were monosemantic but non-task-relevant (e.g. text detectors, edge or corner detectors). Attention maps distinguished monosemantic heads (precise chirp localization) from polysemantic heads (multiple irrelevant regions). These findings revealed functional specialization in ViTs, showing how heads processed relevant vs. extraneous information. By decomposing transformers into interpretable components, this work enhanced model understanding, identified vulnerabilities, and advanced safer, more transparent AI.

## 1. Introduction

The increasing complexity of machine learning (ML) models has raised concerns about their interpretability and safety. Explainability techniques offer a pathway to address these concerns by providing insights into how models make predictions. These techniques can be used for model validation, debugging, and knowledge discovery. From an explainable AI (XAI) perspective, understanding the factors that influence model predictions is essential for ensuring trustworthiness and reliability (Simon et al., 2024).

Explainability techniques can be broadly categorized into two main groups: input-based techniques and model-based techniques (Huang et al., 2024). Input-based techniques focus on manipulating the input data and observing how the model's behavior or predictions change in response. These methods aim to understand the model by analyzing the relationship between input variations and output results. On the other hand, model-based techniques concentrate on modifying or examining the internal structure of the model itself. By probing the model's architecture, parameters, or mechanisms, these techniques provide insights into how the model processes information and makes decisions.

Black-box perturbation methods fall under the category of input-based techniques and involve testing how model predictions change when unexpected inputs are introduced (Ali et al., 2022). One such technique is Permutation Importance, where the values of each feature are shuffled to assess their impact on predictions. If shuffling a feature significantly degrades model performance, it indicates that the model heavily relies on that feature. However, this method may underestimate the importance of highly correlated features.

Another class of input-based techniques includes example-based explanations. Humans often rely on examples to understand complex systems. In XAI, example-based explanations include prototypes and counterfactual explanations. Prototypes (Yampolsky et al., 2024) are representative examples of the data distribution, often identified through clustering algorithms like k-means. Counterfactual explanations identify minimal changes to an input that would alter the model's decision (Baron, 2023).

Local Interpretable Model-agnostic Explanations (LIME) and SHapley Additive exPlanations (SHAP) are two additional techniques that rely on sample inputs to provide interpretable and model-agnostic explanations. LIME approximates a model's prediction for a specific data point by perturbing the input and training a linear model on the weighted dataset (Ribeiro et al., 2016). This technique highlights the features that contribute most to a prediction. SHAP values quantify the marginal contribution of each feature to the model's output. By building local approximations of the model near each data point, SHAP provides insights into feature importance (Tarabanis et al., 2023).

Another input-based technique is occlusion which involves masking parts of the input and observing changes in model activation. For image data, a sliding patch is used to occlude different regions, generating a heatmap of the most sensitive areas (Resta et al., 2021).

Adversarial examples represent another class of input-based techniques (Chanda et al., 2025), which involve making minimal changes to an input to fool the model. These perturbations are often imperceptible to humans but can significantly alter model predictions. Techniques like the Fast Gradient Sign Method (FGSM) optimize inputs to maximize prediction error, revealing vulnerabilities in the model (Card et al., 2024).

While explainability techniques have made significant contributions by revealing how input features influence model predictions, they often fall short of explaining how specific attributes—such as size, color, pattern, texture, direction, and shape—impact the model's decision-making process. Additionally, these techniques do not fully elucidate how intermediate computational structures within the model respond to these attributes. Mechanistic interpretability, however, offers a promising approach to addressing these limitations by providing deeper insights into the internal mechanisms of the model (Pîslar et al., 2025).

Mechanistic Interpretability is an emerging subfield that provides a structured approach to understanding how complex models make decisions (Rai et al., 2024). It focuses on breaking down neural networks into smaller, interpretable components. In a network, neurons can be monosemantic (responding to a single type of input) or polysemantic (responding to multiple unrelated inputs). Polysemantic neurons emerge due to superposition, where different features are shared across a limited number of neurons. Monosemantic neurons can also form circuits—subgraphs within the network that represent meaningful computational pathways. By studying these circuits, we gain deeper insights into how individual features interact and contribute to the model's overall behavior (Olah et al., 2020). Branch specialization is another emerging behaviour of vision transformers where layers can split into branches that specialize in specific features. This specialization mirrors another form of functional organization of biological systems, such as the brain (Voss et al., 2021).

A key element of large models, such as transformers, is the attention mechanism, with attention heads serving as its fundamental building blocks (Elhage et al., 2021). These heads play an important role in facilitating the flow of information between tokens, enabling the model to capture complex dependencies and relationships within the input data. In multi-head attention, data is projected into multiple spaces, with each head learning its own transformations. This allows the model to capture diverse relationships in the data, such as structured patterns like lines, edges, or curves. For instance, one head might focus on patches forming a straight line, while another might detect curved shapes. However, attention heads can also latch onto spurious cues—irrelevant patterns in the data that lead to biased or unreliable decisions (O'Mahony et al., 2024). For example, a model might mistakenly associate background textures in images with its predictions, compromising generalizability and fairness. Identifying and mitigating such vulnerabilities is critical for AI safety. By analyzing which heads focus on irrelevant features, we can pinpoint weaknesses in the model's learning process and improve its robustness.

Each attention head specializes in capturing specific features, and understanding these roles is key to explaining model decisions (Gandelsman et al., 2023). If certain heads contribute to undesirable biases, they can be removed or replaced to enhance fairness and accuracy. Conversely, high-performing heads can be isolated and preserved. This process of modifying or disabling heads is known as ablation. Ablating heads that are critical to the model's performance significantly degrades its output, while removing less important heads can sometimes improve efficiency. A related technique, activation patching, involves altering specific heads or layers to observe their impact on predictions (Heimersheim et al., 2024). For example, replacing a head associated with spurious cues can flip the model's decision, providing insights into its decision-making process. By intervening—such as zeroing out the output of problematic heads—we can analyze their contributions and refine the model. This not only enhances interpretability but also helps debug and improve the model's performance, ensuring it relies on meaningful features rather than spurious correlations.

In this study, I explore the roles of individual attention heads in vision transformers fine-tuned on distorted 2D representations containing non-relevant elements like axis labels, titles, and color bars. By intentionally incorporating these extraneous features, I analyze how intermediate transformer components respond to unrelated information.

## 2. Method

Figure 1 illustrates the process of extracting and normalizing attention maps from a Vision Transformer (ViT) model, followed by an ablation analysis to evaluate the contribution of individual attention heads and layers. The process begins with input dataset preprocessing (100k images), patch embedding, and ViT forward pass, culminating in attention map extraction and overlay on original images. The ablation analysis involves iterating over layers and heads, zeroing out weights, evaluating the model, and computing loss to assess the impact of specific attention mechanisms.

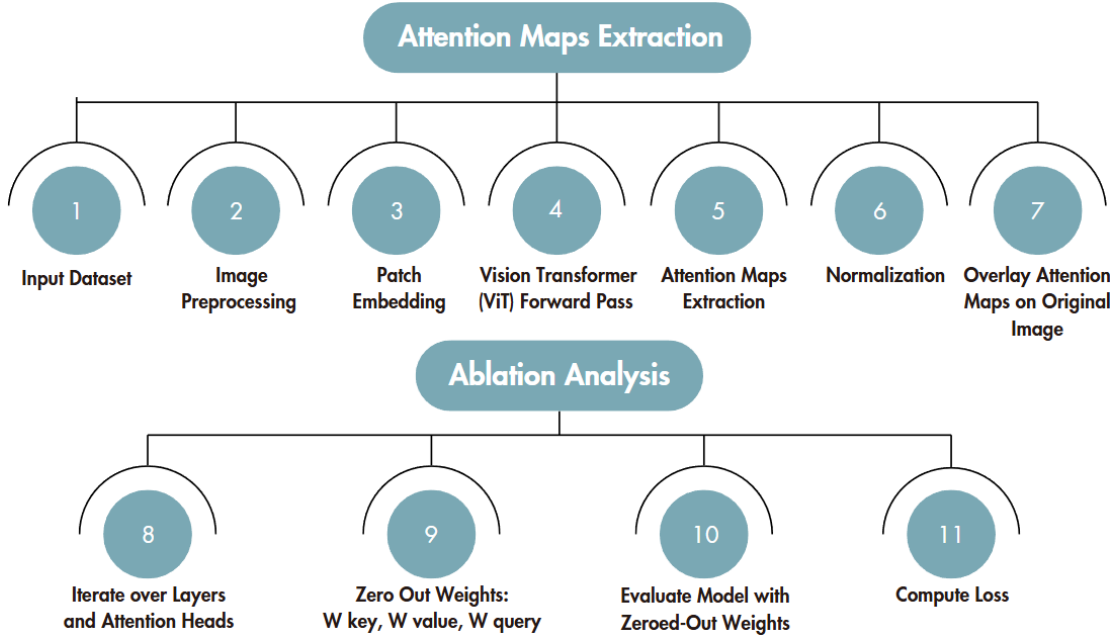

Figure 1: Flowchart of Attention Maps Extraction and Ablation Analysis

### 2.1. Attention Maps Extraction

By leveraging a pre-trained Vision Transformer (ViT) and fine-tuned regression head, the method extracts, normalizes, and visualizes multi-head self-attention maps from input images, providing insights into the model's attention patterns and enhancing interpretability of its decision-making process.

The algorithm begins with a dataset of syntactic images $\chi = \{x_1, x_2, ..., x_N\}$, where each image $x_i \in \mathbb{R}^{H \times W \times C}$ has height $H$, width $W$, and $C$ channels. The goal is to extract attention maps from a pre-trained ViT model $f_\theta$, parameterized by $\theta$, which consists of $L$ layers and $H$ attention heads per layer. A fine-tuned regression head $g_\phi$, parameterized by $\phi$, is used to predict target outputs $y_i \in \mathbb{R}^3$. The algorithm outputs a set of attention maps $A = \{A_1, A_2, ..., A_N\}$, where each $A_i \in \mathbb{R}^{L \times H \times P \times P}$ represents the attention weights for $L$ layers, $H$ heads, and $P \times P$ patches.

The algorithm begins with image preprocessing. Each image $x_i$ is resized to a fixed size $(H', W') = (224, 224)$ using the transformation $x'_i = Resize(x_i, (H', W'))$. The resized image $x'_i$ is then converted into a tensor $\tilde{x}_i \in \mathbb{R}^{C \times H' \times W'}$ through the operation $\tilde{x}_i = ToTensor(x'_i)$.

Next, the algorithm performs patch embedding. The tensor $\tilde{x}_i$ is divided into $P \times P$ non-overlapping patches, where each patch is flattened into a vector $p_{i,j} \in \mathbb{R}^D$, where these patches are projected into a $D-$dimensional embedding space using a learned projection matrix $E \in \mathbb{R}^{D \times D}$, resulting in patch embeddings $z_{i,j} = E \cdot p_{i,j}$. A learnable positional embedding $pos \in \mathbb{R}^{(p^2+1) \times D}$ is then added to the patch embeddings, yielding the final embeddings $Z_i = [z_{cls}, z_{i,1}, z_{i,2}, \ldots, z_{i,P^2}] + pos$, where $z_{cls}$ is the classification token.

During the forward pass through the ViT, the multi-head self-attention (MSA) weights are computed for each layer $l = 1, 2, \ldots, L$. The MSA operation is defined as $MSA(Z_i^{(l)}) = Concat(head_1, head_2, \ldots, head_H) \cdot W^O$, where each head is computed as $head_h$. Here, $Q_h = Z_i^{(l)} W_h^Q$, $K_h = Z_i^{(l)} W_h^K$ and $V_h = Z_i^{(l)} W_h^K$ are the query, key, and value matrices, respectively. The attention weights $A_{i,l,h} \in \mathbb{R}^{(P^2+1) \times (P^2+1)}$ for each $head_h$ are extracted, and the patch embeddings $Z_i^{(l+1)}$ are updated using the MSA output and a feed-forward network (FFN) as $Z_i^{(l+1)} = FFN\left(LayerNorm\left(Z_i^{(l)} + MSA(Z_i^{(l)})\right)\right)$.

To extract attention maps, the algorithm processes each image $x_i$ to obtain the attention weights $A_{i,l,h}$ for all layers $l$ and heads $h$. The attention weights corresponding to the classification token are excluded, resulting in $A_{i,l,h} = A_{i,l,h}[1:, 1:]$. These weights are then reshaped to match the patch grid $P \times P$, yielding $A_{i,l,h} \in \mathbb{R}^{P \times P}$.

For normalization, the attention weights $A_{i,l,h}$ are scaled to prepare them for visualization. The normalized attention weights $\tilde{A}_{i,l,h}$ are computed as $\tilde{A}_{i,l,h} = \frac{A_{i,l,h} - min(A_{i,l,h})}{max(A_{i,l,h}) - min(A_{i,l,h})}$.

To overlay the attention maps on the original image, the algorithm computes the attention contribution $\tilde{A}_{i,l,h}(m, n)$ for each patch $(m, n)$ in the patch grid. The attention weights are overlaid on the original image $x_i$ using the operation, where $\alpha$ is a scaling factor for visualization.

Finally, the attention maps $\tilde{A}_{i,l,h}$ and overlaid images are saved for further analysis.

### 2.2. Ablation Analysis

The ablation analysis systematically evaluates the contribution of each attention head and layer to the model's performance by zeroing out the weights of specific attention heads and measuring the resulting impact on the model's predictions. Performance is quantified using Mean Squared Error (MSE) loss. Followings are the details: The model was adapted for regression tasks using a pretrained ViT backbone with Low-Rank Adaptation (LoRA), followed by a regression head. The ViT backbone processes input spectrogram images $x \in \mathbb{R}^{H \times W \times C}$, where $H = W = 224$ and $C = 3$, to produce a hidden state representation $h \in \mathbb{R}^d$, with $d$ being the hidden size of the ViT. LoRA introduces low-rank matrices $A \in \mathbb{R}^{r \times d}$ and $B \in \mathbb{R}^{d \times r}$ to adapt the weights of specific layers, such as the $query$ and $value$ layers. The adapted weight $W'$ is computed as $W' = W + \alpha \cdot (A \cdot B)$, where $W$ is the original weight matrix, $r$ is the rank of the adaptation, and $\alpha$ is a scaling factor. The regression head, a fully connected neural network, takes the CLS token representation $h_{CLS} \in \mathbb{R}^d$ and produces regression predictions $y \in \mathbb{R}^3$ through a series of linear transformations and ReLU activations: $y = W_2 \cdot ReLU(W_1 \cdot ReLU(W_0 \cdot h_{CLS} + b_0) + b_1) + b_2$, where $W_0, W_1, W_2$ are weight matrices and $b_0, b_1, b_2$ are biases. The dataset was preprocessed by resizing each spectrogram image to $224 \times 224$ and normalizing it using mean $\mu$ and standard deviation $\sigma$. The preprocessed images and their corresponding labels were loaded into a DataLoader with a batch size of 32 for efficient evaluation. The evaluation process involves computing the Mean Squared Error (MSE) loss $\mathcal{L} = \frac{1}{N}\sum_{i=1}^{N}\|y_i - \hat{y}_i\|^2$, where $y_i$ is the ground truth, $\hat{y}_i$ is the predicted output, and $N$ is the batch size. To perform ablation analysis, the algorithm iterates over each layer $l \in \{0,1,...,L-1\}$ and each attention head $h \in \{0,1,...,H-1\}$ in the ViT model. For each layer-head combination, the weights of the specified attention head are zeroed out: $W_{key}^{(l,h)} = 0, W_{value}^{(l,h)} = 0 \text{ and } W_{query}^{(l,h)} = 0$. The model is then evaluated, and the loss $\mathcal{L}^{(l,h)}$ is computed.

## 3. Results

To analyze the behavior of attention heads in detecting chirp pattern, synthetic spectrograms with diverse chirp characteristics were used. The visualization of 15 samples in Figure 2 highlights the diversity in chirp characteristics, including their temporal and spectral properties, as well as the impact of background noise on their spectrogram representation.

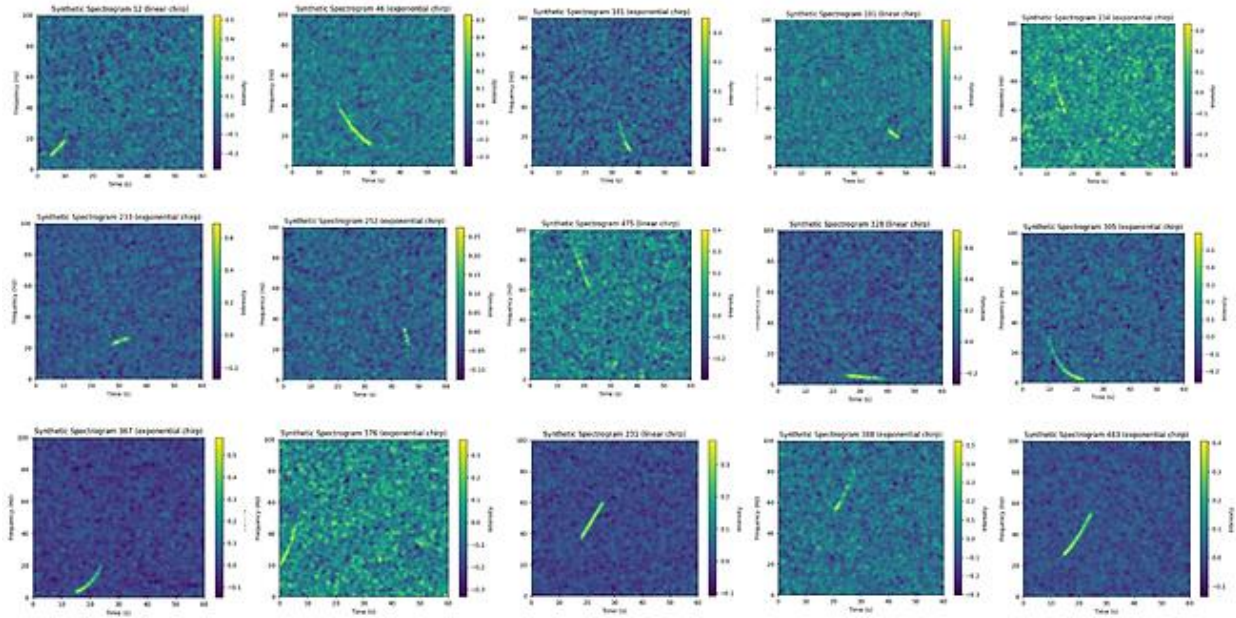

**Figure 2**: Visualization of synthetic spectrogram samples featuring chirps at different positions within the spectrogram, varying in duration, frequency bands, type (linear/nonlinear), and background noise levels.

The Figure 3 illustrates the effect of ablating individual attention heads in a ViT model fine-tuned for a regression task. It presents a 3x4 grid of histograms, where each subplot corresponds to a specific layer (1 to 12) and displays the distribution of raw predictions for chirp start time across all 12 heads within that layer. The baseline prediction distribution from the unablated model (shown in gray) is overlaid in each subplot for comparison. This visualization highlights how ablating specific heads impacts the model's predictions relative to the baseline.

One observation from the Figure 3 is that ablating attention heads in the first three layers has minimal impact on the model's performance. This is evident from the fact that the prediction distributions for these layers closely resemble the baseline distribution of the unablated model. This suggests that the initial layers may play a less critical role in the regression task, potentially focusing on low-level features that do not significantly influence the final predictions.

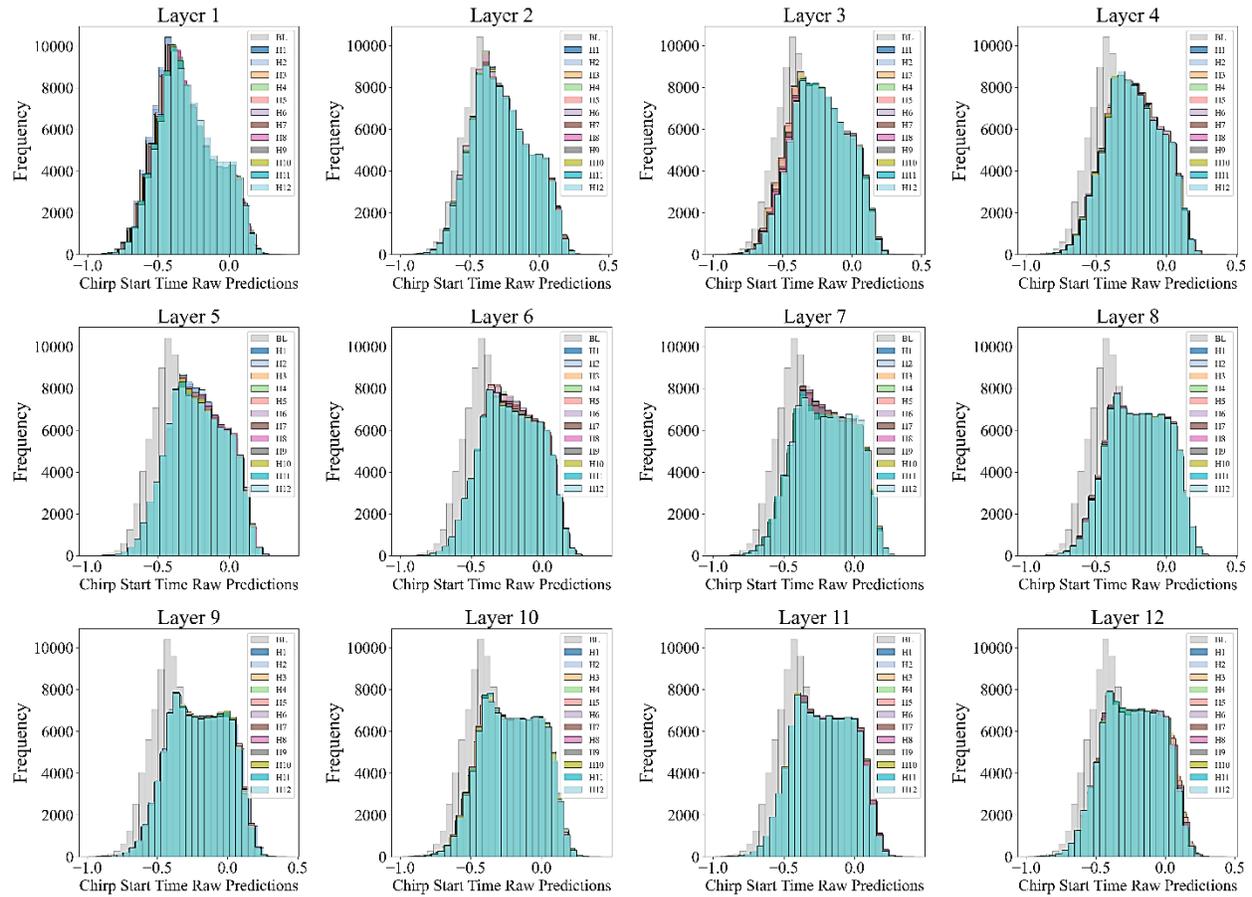

**Figure 3**: A 3x4 grid of histograms comparing the distribution of chirp start time predictions across 12 layers and their respective 12 heads in a neural network model, with a baseline distribution (gray) overlaid for reference. Each subplot represents a specific layer, with 12 distinct colors indicating individual heads. The baseline (BL) represents the model's performance with all heads active, while the other distributions reflect the impact of ablating specific heads.

For ablation analysis, the loss was computed as the Mean Squared Error (MSE) between the model's predictions and the ground truth labels, calculated for each batch during evaluation, and averaged across the entire dataset. The computed loss values were used to analyze the impact of ablating specific attention heads in the model. To visualize how ablating (zeroing out) individual attention heads across different layers affects the MSE loss, a heatmap was generated. The Figure 4 presents this heatmap. The change in MSE loss was measured relative to the baseline loss from the unablated model. Each cell in the heatmap represents the percentage increase in loss when a specific head in a given layer is ablated. Warmer colors

(red) indicate a significant increase in loss, suggesting that the ablated head plays a critical role in the model's performance. Conversely, cooler colors (blue) indicate minimal impact, implying that the corresponding head contributes less to the task.

The heatmap provides strong evidence that the heads in the first three layers have minimal impact on the model's final predictions, as ablating them results in the smallest increase in loss. This is indicated by cooler colors (e.g., blue) in the heatmap. A small increase in loss after ablation suggests that the ablated head was not critical to the model's performance. In other words, removing these heads does not significantly degrade the model's ability to make accurate predictions. This suggests that the heads in the first three layers are either: 1) Redundant—other heads can compensate for their absence. 2) Focused on less important features that do not strongly influence the final predictions. This is likely because, in Vision Transformers (ViTs), the early layers typically focus on extracting low-level features from the input, such as edges, textures, or simple patterns. These low-level features are often less task-specific and may not directly contribute to the final regression task (e.g., predicting chirp start time, start frequency, and end frequency). Therefore, if computational efficiency is a concern, the model could potentially be optimized by reducing the number of heads in the early layers without significantly impacting performance.

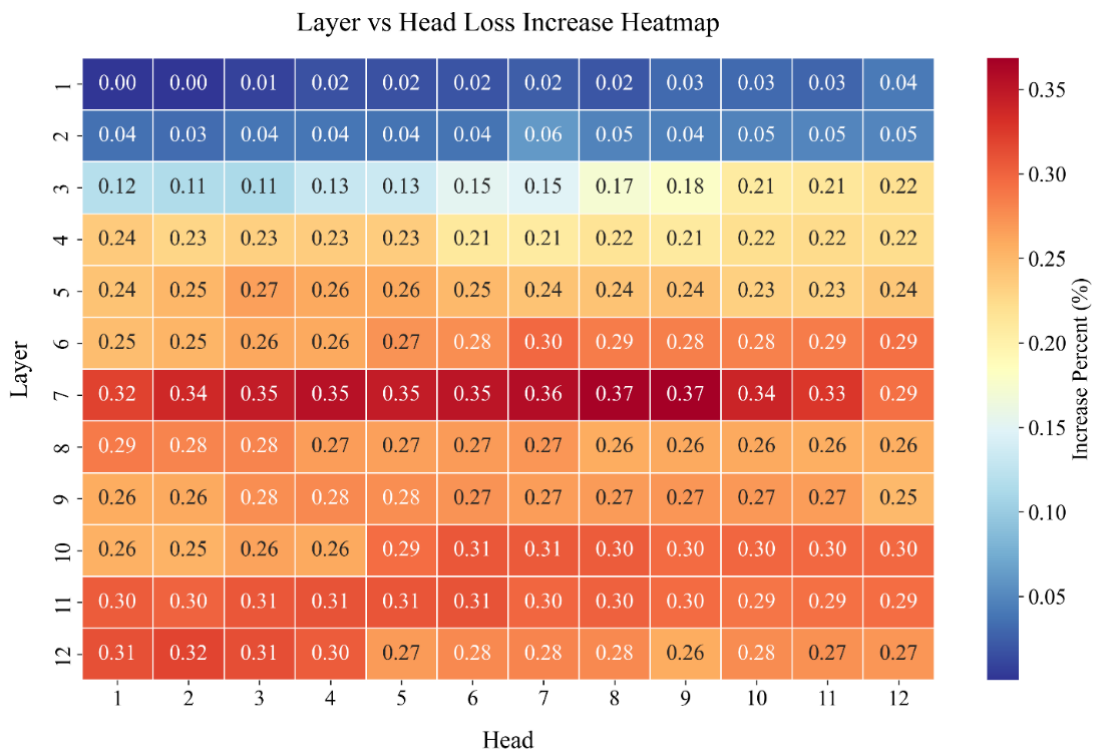

**Figure 4**: The heatmap shows the percentage increase in MSE loss after ablating individual attention heads across layers in the fine-tuned ViT model, compared to the loss of the unablated model. Each cell represents the percentage increase in loss for a specific layer-head combination, with warmer colors (red) indicating higher increases and cooler colors (blue) representing minimal changes.

For each batch of images, the attention weights were retrieved, which provided the attention matrices for all layers and heads. These attention weights were processed to exclude the CLS token and focus on the patch-level attention. The attention weights were averaged across patches and normalized for visualization. Each patch-averaged attention map was overlaid on the original image. This allowed to observe how different heads in different layers attend to various regions of the spectrogram images. Based on the inspection of the attention maps, as seen in the Figure 5, only specific heads in layers 4, 6, 7, 8, 9, 10, and 11 exhibit monosemantic behavior—meaning they solely attend to the pixels where the chirp is located in the spectrogram. These heads demonstrate a focused response to the chirp region, ignoring other non-relevant areas of the spectrogram. In contrast, the remaining heads either attend to the chirp alongside other irrelevant regions or completely fail to capture the chirp, instead focusing on non-relevant parts of the spectrogram. This suggests that only these identified heads in the specified layers are truly monosemantic, as they exclusively respond to the chirp, highlighting their specialized role in the model's ability to localize the chirp pattern.

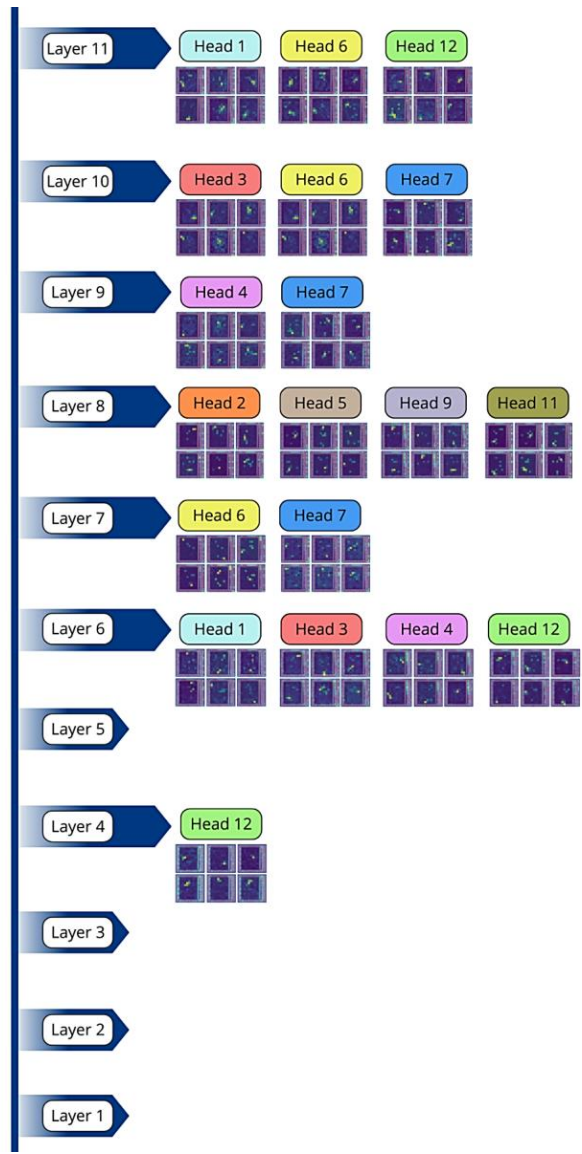

**Figure 5**: Visualization of monosemantic attention heads that focus solely on the chirp location in the spectrogram while disregarding irrelevant regions. These attention heads are specialized in localizing chirp patterns.

Figure 6 shows one of these attention maps that are specialized in localizing chirp patterns. In contrast to the monosemantic heads, Figure 7 demonstrates the attention map of a polysemantic head (Layer 12, Head 12), which responds to multiple irrelevant features across the spectrogram. The attention weights are distributed across non-relevant regions, indicating that this head does not specialize in localizing the chirp but instead captures a variety of unrelated patterns.

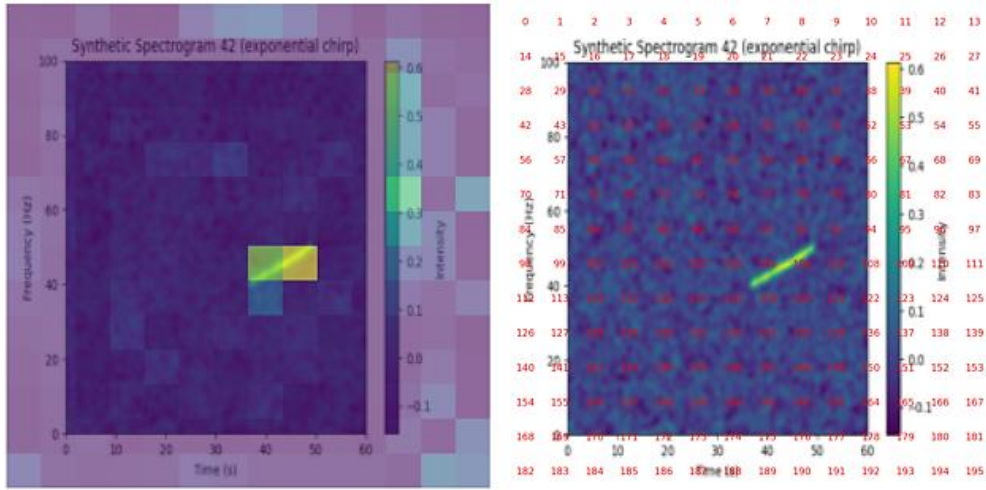

**Figure 6**: Attention Head (Layer 6, Head 4) with Monosemantic Behavior: This figure shows attention map from one of the heads specialized in localizing chirp patterns. The attention weights are focused precisely on the region of the spectrogram where the chirp is located, demonstrating monosemantic behavior.

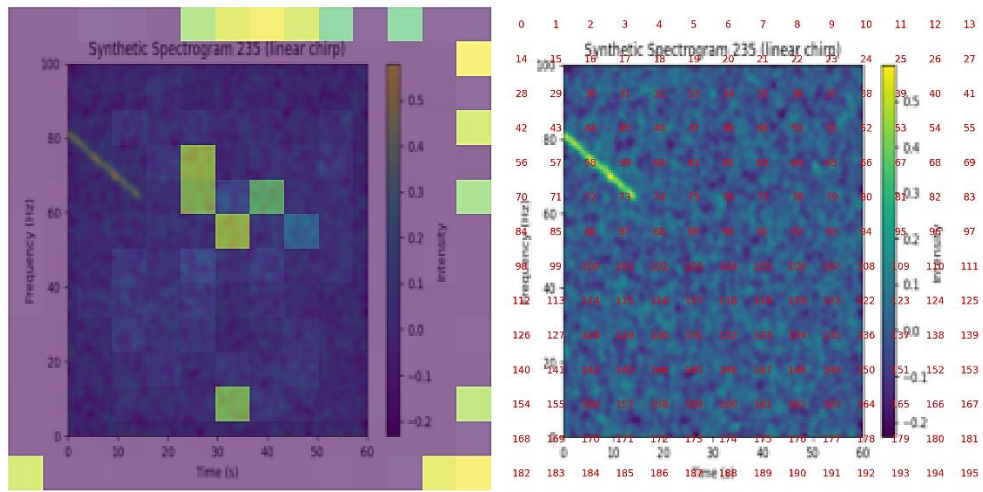

**Figure 7**: Attention Head (Layer 12, Head 12) with Polysemantic Behavior: This figure illustrates the attention map of a polysemantic head. Unlike monosemantic heads, this head attends to multiple regions of the spectrogram, including various irrelevant features. The diffuse attention pattern highlights the head's lack of specialization, as it responds to a broader range of features rather than focusing exclusively on the chirp signal.

Several monosemantic attention heads were observed in the first four layers of the model (Figure 8), specializing in detecting specific visual features beyond the chirp pattern. For instance, Head 1 in Layer 1 functions as a text detector, identifying axis labels, numeric text, and figure titles. Head 5 in Layer 1 is dedicated to detecting color bars. In Layer 2, Head 1 detects the upper-left inner corner, Head 4 identifies the left edge, Head 5 focuses on the top edge, and Head 12 captures the right edge. Moving to Layer 3, Head 4 detects the top corner, Head 7 identifies the bottom outer left corner, and Head 11 specializes in the left inner corner. In Layer 4, Head 2 and 3 respectively detect the lower and upper parts of the color bar, Head 5 identifies bottom axis labels, and Head 8 focuses on color bar labels. These heads in the first four layers are monosemantic but specialize in detecting features that are not directly relevant to the primary task.

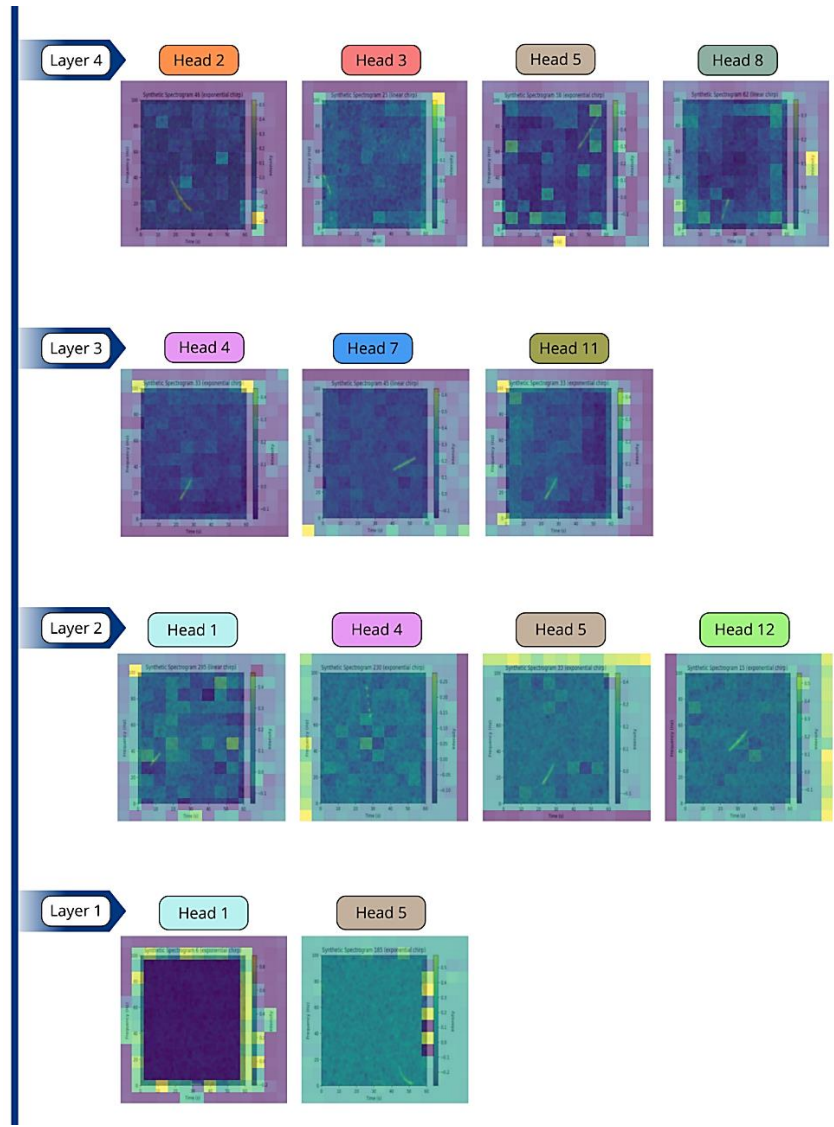

**Figure 8**: Visualization of monosemantic attention heads with non-task-relevant functions across the first four layers: Examples include text detection (axis labels, numeric text, titles), color bar identification, and edge/corner detection (e.g., top edge, left inner corner).

## 4. Discussion

The primary objective of most scientific research is to uncover and understand the causal mechanisms that govern the relationships between variables. To achieve this, structural causal models (SCMs) have emerged as a powerful framework for representing such mechanisms. SCMs provide a formal foundation for addressing causal questions across all three levels of Pearl's causal hierarchy: observational, which involves understanding associations in the data; interventional, which explores the effects of actions or interventions; and counterfactual, which examines hypothetical scenarios and alternative outcomes (Sick et al., 2025). This highlights the importance of understanding the reasons behind specific relationships or outcomes. Explainability and interpretability techniques can enhance causal understanding by revealing how inputs and model structures influence outputs.

Input-based explainability techniques focus on analyzing and manipulating the input space to interpret and explain the behavior of the model's outputs. One example approach leverages Shapley values to evaluate the contribution of individual attributes to local model training, providing a granular understanding of feature importance (Li et al., 2025). Similarly, masking the contribution of specific components to classification performance has emerged as a powerful tool to isolate and analyze the role of individual elements within a model (Xompero et al., 2025). These methods enhance the ability to interpret model behavior by input manipulation. Model-based explainability techniques, on the other hand, aim to dissect and understand the inner workings of complex models, with mechanistic interpretability emerging as a promising subfield within this area. Mechanistic interpretability aims to improve the safety and ethical alignment of AI systems.

One area of research focused on the encoding of toxicity levels in language models (LMs). Studies have shown that LMs strongly encode information about the toxicity of inputs, particularly in their lower layers (Waldis et al., 2025). The use of sparse autoencoders (SAEs) has also gained traction, particularly in the context of biological mechanisms. By training SAEs on the residual stream of models, researchers can generate hypotheses about underlying biological processes (Adams et al., 2025). Furthermore, mechanistic interpretability has proven invaluable in identifying flawed heuristics within models as well as designing more robust synthetic datasets (Quirke et al., 2025). This dual application of interpretability—both in understanding and improving models—highlights its versatility. Additionally, techniques such as edge attribution patching and sparse autoencoders have been used to identify minimal circuits and components within models, providing insights into the fundamental building blocks of model behavior (Quirke et al., 2025). These methods are particularly useful for distilling complex models into more interpretable sub-components.

Concept polysemanticity disentanglement represents another frontier in mechanistic interpretability. By filtering and distinguishing the most contextually relevant concept atoms, researchers can reduce ambiguity in model interpretations. The introduction of Concept Polysemanticity Entropy (CPE) as a quantitative measure of concept uncertainty further refines this approach, enabling the modeling of uncertain concept atom distributions (Yu et al., 2025). This advancement not only enhances interpretability but also provides a framework for quantifying the reliability of model explanations.

Attention mechanisms have also been a focal point of interpretability research. For instance, diversity loss has been proposed to encourage variation among attention heads, ensuring that each head captures distinct spatial representations (Peled et al., 2025). Moreover, emergent structures within these models have been shown to approximate debiased gradient descent predictors (He et al., 2025). The interplay between the number of attention heads and tasks has also been explored, revealing a superposition phenomenon that efficiently resolves multi-task in-context learning (He et al., 2025; Olsson et al., 2022). These findings provide a deeper understanding of how attention mechanisms contribute to model understanding and performance.

In the realm of topic interpretability, innovative approaches such as representing topics as Multivariate Gaussian Distributions and documents as Gaussian Mixture Models have been proposed. This probabilistic framework enhances the interpretability of topic models by providing a more nuanced representation of document-topic relationships (Sahoo et al., 2025). The development of the Dynamic Bidirectional Elman with Attention Network (DBEAN) integrates bidirectional temporal modeling with self-attention mechanisms, dynamically assigning weights to critical input segments. This not only improves contextual representation but also maintains computational efficiency, making it a promising tool for interpretable AI (Lai et al., 2025).

Discovering human-explainable features in clinical tasks is another pathway toward more interpretable and trustworthy AI in healthcare (Chen et al., 2025). The future of mechanistic interpretability depends on refining these techniques and integrating them into AI systems, paving the way for more ethical, transparent, and trustworthy AI development while enhancing accountability in its applications.